\documentclass[a4paper,biblatex]{jacow}
\usepackage{pdfpages,multirow,ragged2e,hyperref} %
\makeatletter%
	\ifboolexpr{bool{xetex}}
	 {\renewcommand{\Gin@extensions}{.pdf,%
	 .png,.jpg,.bmp,.pict,.tif,.psd,.mac,.sga,.tga,.gif,%
	 .eps,.ps,}}{}
\makeatother

\ifboolexpr{bool{xetex} or bool{luatex}} 
{}{\usepackage[utf8]{inputenc}}
\usepackage[USenglish]{babel}

\ifboolexpr{bool{jacowbiblatex}}%
{%
 \addbibresource{TUPD113.bib}
}{}
\newcommand{\code}[1]{\texttt{#1}}
\begin{document}

\title{Unsupervised Anomaly Detection in ALS EPICS Event Logs}

\author{A.~Sulc\thanks{asulc@lbl.gov}, T.~Hellert, S.~Hunt, LBNL, Berkeley, CA, USA}
	
\maketitle

\begin{abstract}
This paper introduces an automated fault analysis framework for the Advanced Light Source (ALS) that processes real-time event logs from its EPICS control system. By treating log entries as natural language, we transform them into contextual vector representations using semantic embedding techniques. A sequence-aware neural network, trained on normal operational data, assigns a real-time anomaly score to each event. This method flags deviations from baseline behavior, enabling operators to rapidly identify the critical event sequences that precede complex system failures.
\end{abstract}

\section{INTRODUCTION}
The Advanced Light Source (ALS) at Lawrence Berkeley National Laboratory is a third-generation synchrotron light source facility serving a diverse international community of scientific users~\cite{Hellert:2024zrj}. Its operation relies on a complex, distributed control system based on the Experimental Physics and Industrial Control System (EPICS) toolkit~\cite{DALESIO1994179}. This system comprises hundreds of Input/Output Controllers (IOCs) that manage over 200,000 process variables (PVs), each corresponding to a specific signal for monitoring or control—from magnet power supplies and vacuum gauges to insertion devices and beamline optical components.

Maintaining high availability and operational efficiency in such a complex environment necessitates rapid fault diagnosis. When the electron beam is lost or a critical subsystem fails, operators face the challenge of identifying the root cause amidst a flood of alarms and status changes that often precede or accompany the actual fault. A single fault can trigger a cascade of secondary events, generating thousands of log entries within seconds and obscuring the initial cause. The ALS employs an "Event Logger" system, which chronologically records state changes for thousands of critical PVs. While this logger provides an invaluable, high-fidelity record of system activity, its sheer volume and unstructured, text-based nature make manual post-mortem analysis time-consuming and heavily reliant on the accumulated expertise of senior staff.

Conventional monitoring often relies on predefined limits and alarms set on individual PVs. Although essential, this approach may not capture subtle precursor events or faults arising from complex, emergent interactions between subsystems. This paper proposes a machine learning framework designed to overcome these limitations. We reframe the problem by treating the Event Logger data as a sequential language and apply modern natural language processing (NLP) and deep learning techniques to learn the "grammar" of normal machine operation. Our system automatically detects and scores anomalous event sequences, providing a quantitative measure of deviation from normal operation for control room operators.

This work aims to shift the diagnostic paradigm from reactive, alarm-based monitoring to a proactive, data-driven approach, with two main contributions: (i) the application of NLP embeddings to create meaningful numerical representations of EPICS events for visualizing and interpreting the machine state; and (ii) a sequence-aware, one-class neural network model to detect anomalies in these event streams.

\section{METHODOLOGY}
Our framework consists of two main stages: (1) data preprocessing and representation, where raw event logs are converted into numerical vectors, and (2) anomaly detection, where a deep learning model scores events based on learned normal patterns.

\subsection{Channel Name Interpretation}
To better understand the inherent structure of the PV naming convention at the ALS, we developed a method to represent the relationships among channel name components as a graph. Although the EPICS naming scheme is mostly hierarchical, this structure can be obscured when viewing thousands of individual PVs. Our visualization technique helps reveal the underlying "grammar" of the control system's nomenclature, confirming that channel names are both parseable and scalable.

\subsubsection{ALS EPICS Channel Name Convention}
The established naming convention follows the structure \code{SysSubSys:DeviceID:Signal}, with an optional \code{SubDevice} component included as \code{SysSubSys:DeviceId-SubDevice:Signal}. This format is designed for programmatic parsing, with the colon (\code{:}) as a reserved delimiter. The segment between the first and second colons constitutes the \code{DeviceID}, and everything following the second colon is the \code{Signal} name. To ensure consistency, \code{Sys} (System) and \code{Device} names are restricted to an agreed-upon list and cannot contain colons. Furthermore, device names must not end in a number to avoid ambiguity with the \code{DI} (Device Instance), which must be a number without leading zeros.
\begin{figure*}[ht!]
\centering
\includegraphics[width=1.0\linewidth]{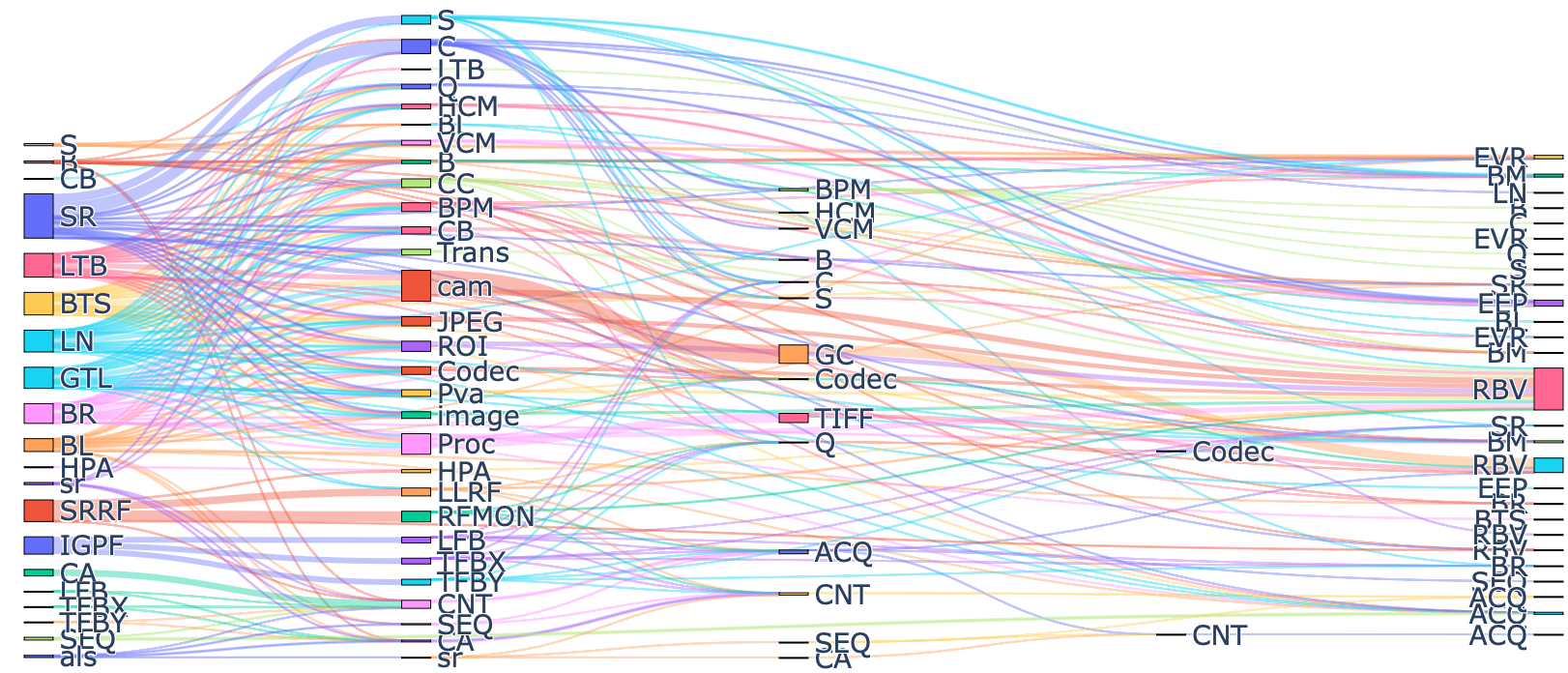}
\caption{Sankey diagram illustrating the hierarchical structure of the ALS EPICS channel naming convention. Each node represents a token from a PV name, split by delimiters like colons, underscores, or numbers. The flow from left to right shows common sequences in which these tokens appear. The thickness of each path is proportional to the frequency of that specific sequence across all analyzed PVs. This visualization reveals the underlying connections between individual tokens and the organization of the control system.}
\label{fig:sankey}
\end{figure*}

Several rules govern channel name components. Names are case-sensitive, and capitalization is encouraged for readability. The use of underscores should be minimized, though a signal name ending with an underscore (\code{\_}) specifically denotes a private signal within an IOC. While the \code{SubSys} and \code{DI} fields are not always required, using a \code{DI} of 1 is recommended if more devices of the same type might be added later. The maximum length of a channel name is still under discussion, but developers should avoid characters such as \code{\{\}, [], ., -, +, =, \~, *, |, <, >}.

The process begins by treating each tokenized channel name as a path. For instance, a PV name tokenized into \code{SR}, \code{HCM}, and \code{ERR} is interpreted as a directed path from a Storage Ring (\code{SR}) node, through a Horizontal Corrector Magnet (\code{HCM}) node, to an Error (\code{ERR}) node.

\subsection{Data Source and Preprocessing}
The primary data source is the ALS Event Logger, which produces text files of tab-separated entries. Each entry includes a timestamp (with millisecond resolution), the PV name, its previous and new states, and an explanation string (e.g., from the EPICS record's \code{DESC} field).

Before analysis, the raw logs undergo several critical preprocessing steps. First, a curated list of known noisy or operationally irrelevant PVs is filtered out. These often include signals related to routine user interactions (e.g., video camera controls) or high-frequency, benign status updates (e.g., \code{LNRF:MOD:WarningExists}) that do not contribute to fault diagnosis. Second, the core textual information of each event—the PV name and its explanation string—is concatenated and tokenized. Tokenization breaks down these strings into their constituent parts based on non-alphanumeric characters. For example, the PV name \code{SR07U:GDS1E:BC02} is split into tokens like \code{SR07U}, \code{GDS1E}, and \code{BC02}. This approach allows the model to learn the semantic meaning of individual components within the hierarchical EPICS naming convention without explicit programming.

The entire hierarchy of all channels present in the EPICS control system is shown in Fig.~\ref{fig:sankey}.

\subsection{Event Representation as Vectors}
A fundamental prerequisite for applying machine learning algorithms to textual data is the transformation of symbolic information into a numerical format. The raw, tokenized text from the EPICS event logs must be converted into a meaningful, quantitative representation that a model can process. The effectiveness of any subsequent anomaly detection relies on the quality of this initial representation, as it must accurately capture the rich operational semantics embedded within the channel names and event descriptions.

Our methodology is founded on a powerful analogy: we treat the stream of event logs as a specialized language. In this framework, each individual event log entry is considered a "sentence." The constituent parts of that event—the tokens extracted from the PV name and its associated strings during preprocessing—are treated as the "words" that compose the sentence. This conceptual shift allows us to leverage sophisticated techniques from the field of Natural Language Processing (NLP).

Traditional methods for vectorizing text, such as one-hot encoding, are profoundly inadequate for this task. Such an approach would assign a unique vector to every token, creating two prohibitive problems. First, it leads to extremely high-dimensional and sparse vectors, which are computationally inefficient and susceptible to the "curse of dimensionality." Second, and more critically, this method treats every token as an independent, isolated entity. It completely fails to encode any notion of semantic similarity; for example, the vectors for tokens representing a horizontal corrector magnet and a vertical corrector magnet would be mathematically orthogonal, implying no relationship, which is operationally untrue.

To overcome these limitations, we employ Word2Vec, a predictive deep learning model developed to learn high-quality, distributed vector representations of words from large text corpora~\cite{mikolov2013}. Since Word2Vec is parameter-efficient, we can train it on a smaller corpus compared to many deep learning approaches. We train a Word2Vec model on our event logs that appear within a specified time range. The model learns a dense vector representation, known as an embedding, for each unique token by analyzing the contexts in which it appears. The core principle is that tokens that frequently occur in similar surrounding contexts will be mapped to vectors that are close to one another in the resulting high-dimensional vector space. This geometric proximity becomes a mathematical proxy for semantic relatedness.

\begin{figure}[h]
\centering
\includegraphics[width=1.0\linewidth]{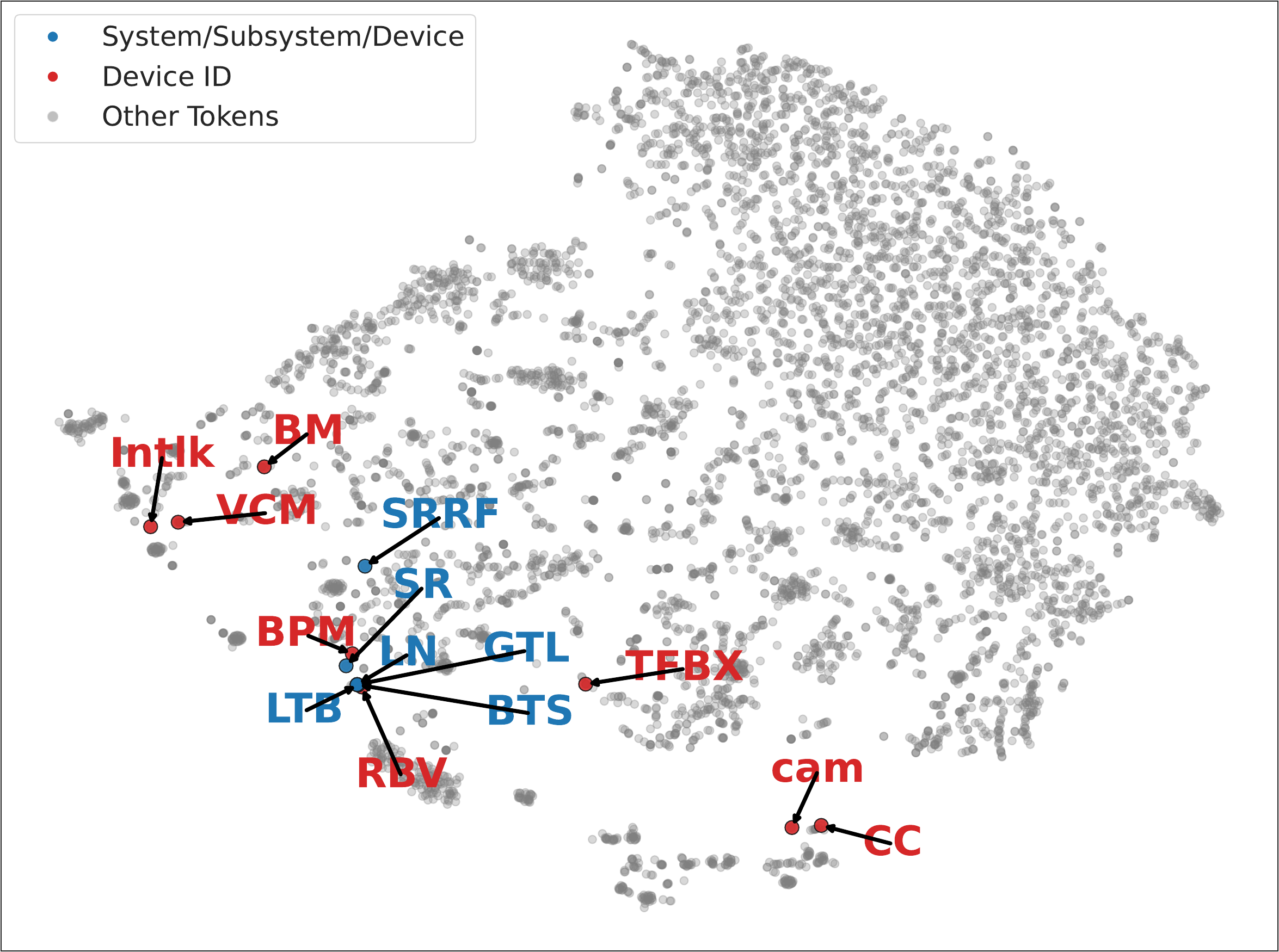}
\caption{t-SNE embedding of individual subparts of PV names in our EPICS system, with \texttt{SysSubSys} Group 1 distinguished by blue and \texttt{DeviceID} Group 2 denoted by red (numbers removed).}
\label{fig:wordvec}
\end{figure}

A key advantage of Word2Vec embeddings is their compositional nature. The vector representation for a multi-token "sentence," such as an entire event, can be effectively constructed by aggregating the vectors of its constituent "word" tokens. For an event composed of the tokens $t_1, t_2, \dots, t_N$, we compute a single vector representation for the event, $\vec{E}$, by summing the individual embedding vectors:
$$ \vec{E} = \sum_{i=1}^{N} \vec{v}(t_i) $$
where $\vec{v}(t_i)$ is the learned embedding vector for the token $t_i$. To provide a concrete example, consider an event associated with the PV \texttt{SR07U:GDS1E:BC02}. After tokenization, we have the tokens \texttt{SR07U}, \texttt{GDS1E}, and \texttt{BC02}. The final vector for this event is calculated by vector addition:
$$ \vec{E}_{\text{SR07U:GDS1E:BC02}} = \vec{v}(\text{SR07U}) + \vec{v}(\text{GDS1E}) + \vec{v}(\text{BC02}) $$
This process translates each discrete log entry into a specific point within a continuous vector space. The structure of this space is not arbitrary; rather, its geometry inherently reflects the operational relationships within the control system. Events involving similar subsystems or actions will naturally lie in proximity to one another. The visualization in Fig.~\ref{fig:wordvec}, which uses t-SNE to project the high-dimensional token embeddings into a two-dimensional plane, provides visual confirmation of this semantic organization. It shows that tokens from the same functional groups form distinct clusters, validating our hypothesis about the underlying grammar of the EPICS naming convention.

Having established a method for vector representation, we can now embed each incoming event in real-time. This transforms the raw text stream into a sequence of meaningful vectors, which is the ideal input for the modern anomaly detection algorithms described in the following section.

\subsection{Sequence-Based One-Class Anomaly Detection}
Following the vectorization of discrete events, the subsequent analytical step is to model the temporal dynamics of the control system. The diagnostic significance of an event is heavily dependent on its context within a sequence. Therefore, our model must be capable of analyzing entire sequences of event vectors rather than isolated data points.

We formulate the problem as \textbf{one-class anomaly detection}~\cite{pmlr-v80-ruff18a}. This methodology is predicated on the observation that the data manifold of nominal operational states is densely populated and can be effectively modeled, whereas the space of anomalous states is vast, sparse, and contains novel failure modes not present in historical data. The objective is to construct a precise model of nominal behavior from an unlabeled training corpus. Any sequence that deviates significantly from this model is classified as an anomaly.

\subsubsection{Deep One-Class Learning Framework}
Our approach utilizes the deep one-class classification framework, which integrates the Support Vector Data Description (SVDD) objective with a deep neural network~\cite{pmlr-v80-ruff18a}. The primary objective is to train a neural network, which functions as a transformation $f_{\theta}$, to map input sequences corresponding to nominal operations into a minimal-volume hypersphere in a learned latent representation space. This hyperspherical boundary encloses the nominal data manifold.

The anomaly score, $s$, for an input sequence $x$ is defined as the Euclidean distance between its latent representation $f_{\theta}(x)$ and the pre-established center $c$ of the hypersphere:
\begin{equation}
s = || f_{\theta}(x) - c ||_2
\end{equation}
The network parameters $\theta$ are optimized exclusively on nominal data to minimize this objective function, thereby compelling the network to learn a compressed representation that captures the salient features of normal operational sequences.

\subsubsection{Model Architecture and Implementation}
The network architecture is specifically designed for sequential data processing and is composed of two primary components:
\begin{enumerate}
    \item \textbf{A Gated Recurrent Unit (GRU) Layer:} The core component for sequence modeling is a GRU~\cite{cho2014}. The GRU iteratively processes the input vector sequence, updating its hidden state at each timestep. This hidden state serves as a distributed representation of the preceding event history. Its internal gating mechanisms, the update and reset gates, regulate the flow of information through the unit. This allows the model to selectively retain relevant past information and discard irrelevant data, which is essential for learning long-range temporal dependencies within the event stream.
    \item \textbf{A Fully Connected (Linear) Layer:} The sequence of hidden states from the GRU is processed by a final linear layer. This layer performs a final transformation, projecting the representation from the GRU's hidden space into the target $z$-dimensional latent space where the hyperspherical boundary is defined.
\end{enumerate}

In accordance with the Deep SVDD methodology~\cite{pmlr-v80-ruff18a}, both the GRU and the linear layers are implemented \textbf{without bias terms}. This architectural constraint acts as a regularizer and is critical for preventing a degenerate solution known as "hypersphere collapse," where the network could learn to map all inputs to a constant value by trivializing its weights.

A significant byproduct of this framework is that the network transformation $f_{\theta}$ provides more than just a scalar anomaly score. The output vector $f_{\theta}(x)$ is itself a rich, \textbf{context-aware embedding} of the input sequence $x$ in the $z$-dimensional latent space. While the initial Word2Vec representation vectorizes a single event in isolation, this new embedding, produced by the GRU, represents an event in the context of its entire preceding sequence. This vector can be interpreted as a compressed summary of the machine's operational state at a specific point in time, which can be leveraged for further analysis, including visualization of the learned feature space.

\section{EXPERIMENTS AND RESULTS}

\subsection{Case Study: Beam Dump Precursor Identification on June 25, 2025}
To demonstrate the model's ability in a real-world scenario, we analyze an authentic event sequence recorded moments before a beam dump on June 25, 2025. The data shows a distinct transition from a nominal operational state to a critical fault condition.

Initially, the event log is populated with routine operational events. For over a minute, the log shows activities like:
\begin{itemize}
\item \texttt{sr07u1:Hor\_mtr\_done:1:0}
\item \texttt{SR06U\_\_\_GDS1PS\_BM00:0:1}
\item \texttt{FE08BL3:PSS111:IsOpen:0:1}
\end{itemize}

During this period, the model consistently assigns very low anomaly scores, with $s$ typically ranging from \textbf{0.012} to \textbf{0.024}. These scores indicate that the sequences are recognized as standard behavior. The network transformation $f_{\theta}$ maps these event sequences to output vectors that lie well within the learned hyperspherical boundary of normal operation.

The state of the system changes abruptly at 11:33:01. The following three events are logged in immediate succession:
\begin{enumerate}
    \item \texttt{SR12S\_\_\_TCUP9\_\_BM:0:1}
    \item \texttt{SR12S\_\_\_TCUP9\_L\_BM:0:1}
    \item \texttt{SR12S\_\_\_UP\_OUT\_BM:0:1}
\end{enumerate}

For this brief sequence, the model calculates dramatically higher anomaly scores of \textbf{1.52}, \textbf{1.07}, and \textbf{1.19}, respectively. This sudden spike, representing an increase by two orders of magnitude, provides a clear and quantitative signal of a significant anomaly.

\begin{figure}
    \centering
    \includegraphics[width=1.0\linewidth]{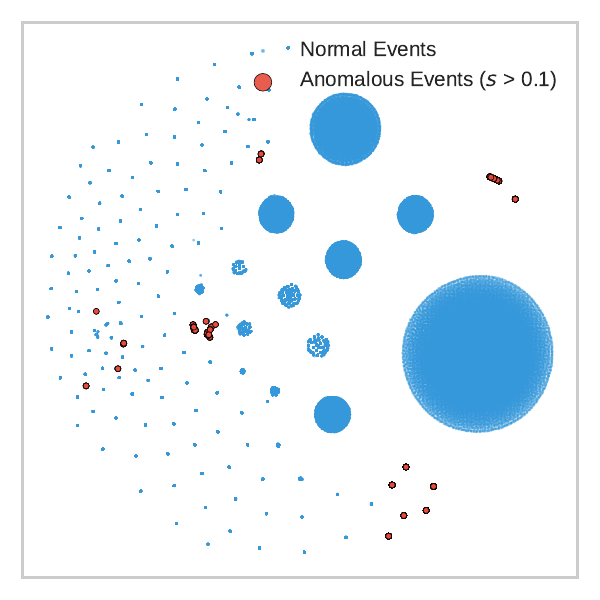}
    \caption{t-SNE visualization of the GRU model's output embeddings ($f_{\theta}(x)$) for events over a two-day period encompassing the beam dump example.}
    \label{fig:embedding}
\end{figure}

This sequence of interlock events, originating from a specific location, represents a severe deviation from any learned pattern of nominal operation. This novel pattern drives the GRU's hidden state into a previously unobserved region of the latent space. As a result, the network maps this subsequence to a point far from the hypersphere's center. Figure~\ref{fig:embedding} shows t-SNE embedding of the events from last two days before the event. 
As result, model successfully filtered out the preceding operational "noise" and automatically pinpointed the critical precursor events immediately preceding the beam dump, correctly identifying the `SR12S` subsystem as the likely origin of the fault.

Note that the event that caused the beam dump was described in our logbook by operators (post mortem) as:
\begin{verbatim}
Beam Loss Root Cause
EPBI SR 12 TCUP9.
We are monitoring SR12 IG 1 temperature in CR.
\end{verbatim}

\subsection{Case Study: Partial Beam Loss on April 11, 2025}
We examined a partial beam loss from April 11, 2025. The operator logbook provided a ground truth for this specific case: \textit{"Lost $\sim$53 mA. Only alarm: \texttt{SR12C\_\_\_QD1\_\_\_\_BM02} the PS On monitor went to 0 at the same time for a moment."} Our goal was to determine if the model's anomaly scores would align with this human diagnosis.

Furthermore, operators also reported in the logbook at 6:40: \textit{ID 7,1 amp trip \texttt{SR07U\_\_\_ODS1PS\_BM04} Vgap Status: 0xD413}

\subsubsection{Analysis of Model Output}
During the period leading up to the fault, the event logs contained many routine updates (e.g., \texttt{sr07u1}). Our model had been trained on similar data and learned these frequent patterns as a signature of normal operation. As a result, it mapped these event sequences to a region well within its learned hypersphere, assigning them consistently low anomaly scores (typically $s \approx 0.04$).

\begin{table}[ht!]
\centering
\caption{Key events and their anomaly scores during the fault.}
\label{tab:partial_loss_tech}
\resizebox{1.0\linewidth}{!}{
\begin{tabular}{l l c c l}
\hline
\textbf{Timestamp} & \textbf{Event Name} & \textbf{State} & \textbf{Score ($s$)} & \textbf{Model Interpretation} \\
\hline
06:36:05 & \texttt{SR07U\_\_\_ODS1PS\_BM04} & \texttt{1:0} & \textbf{0.15} & \textbf{Deviation from normal}\\
\multicolumn{3}{c}{...}\\
06:37:02 & \texttt{SR07U\_\_\_ODS1PS\_BM04} & \texttt{0:1} & \textbf{0.15} & \textbf{Deviation from normal}\\
\multicolumn{3}{c}{...}\\
... & \texttt{sr07u1:Vgap\_mtr\_done} & \texttt{0:1} & 0.04 & Learned normal pattern \\
\textbf{06:38:31.9} & \textbf{\texttt{SR:DCCT5:Ok}} & \texttt{1:0} & \textbf{0.39} & \textbf{Deviation from normal} \\
\textbf{06:38:32.1} & \textbf{\texttt{SR12C\_\_\_QD1\_\_\_\_BM02}} & \texttt{1:0} & \textbf{0.17} & \textbf{Deviation from normal} \\
06:38:33.1 & \texttt{SR12C\_\_\_QD1\_\_\_\_BM02} & \texttt{0:1} & 0.17 & Deviation from normal \\
\hline
\end{tabular}}
\end{table}

\begin{figure}[ht!]
    \centering
    \includegraphics[width=1.0\linewidth]{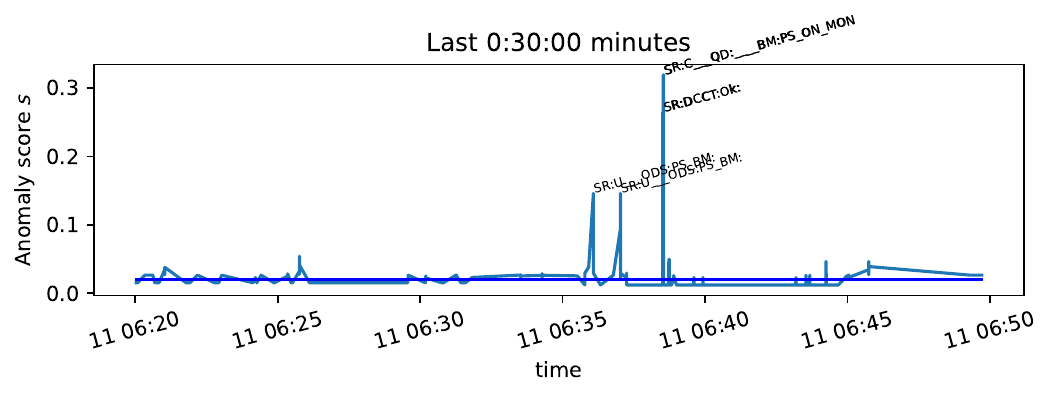}
    \caption{Last 30 minutes scores around the analzed time stamp. Notice two spikes that correpond to event \texttt{SR07U\_\_\_ODS1PS\_BM04} and then sudden spike due to beam dump followed by \texttt{SR12C\_\_\_QD1\_\_\_\_BM02}}
    \label{fig:scores}
\end{figure}

At 06:38:31, the system's state changed. The model processed a short sequence of events that was statistically different from the normal patterns it had learned. This sequence, detailed in Table~\ref{tab:partial_loss_tech}, contained events related to the beam current.

Because this temporal context was novel, the model mapped this new sequence to a point in the latent space far from the center of normal operations. This distance is what produced the elevated anomaly scores:
\begin{itemize}
    \item At 6:40, operators reported a trip of \code{SR07U\_\_\_ODS1PS\_BM04}, which our model also flagged with a score of $s\approx 0.15$.
    \item The beam loss indicator (\texttt{SR:DCCT5:Ok}) received a high score of $s=0.39$.
    \item The power supply monitor (\texttt{SR12C\_\_\_QD1\_\_\_\_BM02}) received a score of $s=0.17$. This value, while lower than that of the beam loss itself, was significantly above the operational background noise.
\end{itemize}

This result is encouraging. It suggests the model learned to distinguish between high-frequency, benign event sequences and rare sequences that are indicative of a fault. The fact that the model assigned a high anomaly score to the exact PV identified by the human operator provides evidence that this data-driven approach could be a useful aid for diagnostics. It appears to have successfully filtered the operational noise and highlighted the event most relevant to the root cause of the partial beam loss.

\section{CONCLUSION}
In this work, we have presented an unsupervised framework for anomaly detection in the EPICS event logs of the Advanced Light Source. By treating log data as a language, we leveraged NLP embedding techniques to create rich, numerical representations of system events. A sequence-aware, deep one-class classification model was then trained to learn the complex grammar of normal machine operations.

Our case studies of real-world beam loss events demonstrate the effectiveness of this approach. The model successfully distinguished routine operational "noise" from critical, fault-related event sequences by assigning significantly higher anomaly scores to the latter. The results show a strong correlation between the highest-scoring events and the root causes identified by human operators, validating our method as a powerful tool for automated diagnostics. This framework represents a significant step beyond traditional, threshold-based alarm systems by providing a context-aware understanding of the machine state.

Future work will focus on the real-time deployment of this system in the ALS control room to provide operators with live diagnostic support. Further research could also explore the use of the learned event embeddings for automated fault classification and predictive maintenance.

\section{Acknowledgement}
We would like to thank our colleagues Thomas Scarvie, Tynan Ford, Simon Leemann, and Hiroshi Nishimura.

This work was supported by the Director of
the Office of Science of the U.S. Department of Energy under Contract No. DEAC02-05CH11231.

\clearpage
{\printbibliography}

\end{document}